\documentclass{article}

\PassOptionsToPackage{numbers, compress}{natbib}
\usepackage[final]{nips_2017}
\usepackage[utf8]{inputenc} 
\usepackage[T1]{fontenc}    
\usepackage{hyperref}       
\usepackage{url}            
\usepackage{booktabs}       
\usepackage{amsfonts}       
\usepackage{nicefrac}       
\usepackage{microtype}      
\usepackage{graphicx}
\usepackage{epstopdf}
\usepackage{amsmath}

\title{Revealing structure components of the retina by deep learning networks}

\author{
	   Qi~Yan, Zhaofei~Yu, Feng~Chen \\
	   Center for Brain-Inspired Computing Research, Department of Automation, Tsinghua University\\
	 \texttt{\{q-yan15,yuzf12\}@mails.tsinghua.edu.cn} \\
	 \texttt{chenfeng@mail.tsinghua.edu.cn} \\
   \And
 Jian K. Liu \\
 Institute for Theoretical Computer Science, Graz University of Technology\\
  \texttt{liu@igi.tugraz.at} \\
}

\begin{document}

\maketitle

\begin{abstract}
Deep convolutional neural networks (CNNs) have demonstrated impressive performance on visual object classification tasks. In addition, it is a useful model for predication of neuronal responses recorded in visual system. However, there is still no clear understanding of what CNNs learn in terms of visual neuronal circuits. Visualizing CNN's features to obtain possible connections to neuronscience underpinnings is not easy due to highly complex circuits from the retina to higher visual cortex. Here we address this issue by focusing on single retinal ganglion cells with a simple model and electrophysiological recordings from salamanders. By training CNNs with white noise images to predicate neural responses, we found that convolutional filters learned in the end are resembling to biological components of the retinal circuit. Features represented by these filters tile the space of conventional receptive field of retinal ganglion cells. These results suggest that CNN could be used to reveal structure components of neuronal circuits.
\end{abstract}

\section{Introduction}

Deep convolutional neural networks (CNNs) have been a powerful model for numerous tasks related to system identification \cite{Lecun2015Deep}. By training a CNN with a large set of targeted images, it can achieve the human-level performance for visual object recognition. However it is still a challenge for understanding the relationship between computation and the underlying structure components learned within CNNs \cite{Smith_2016, Marblestone_2016}. Thus, visualizing and understanding CNNs are not trivial \cite{Zeiler2014Visualizing}.

Inspired by neuroscience studies, a typical CNN model consists of a hierarchical structure of layers \cite{Lecun2010Convolutional}, where one of the most important properties for each convolutional (conv) layer is that one can use a conv filter as a feature detector to extract useful information from inputed images after the previous layer \cite{Simonyan2014Very, Krizhevsky2012ImageNet}. Therefore, after learning, conv filters are meaningful. The features captured by these filters can be represented in the original natural images \cite{Zeiler2014Visualizing}. Often, one typical feature shares some similarities with part of natural images from the training set. These similarities are obtained by using a very large set of specific images. The benefit of this is that features are relative universal for one category of objects, which is good for recognition. However, it also causes the difficulty of visualization or interpretation due to the complex nature of natural images, i.e., the complex statistical structures of natural images \cite{Simoncelli2001Natural}. As a result, the filters and features learned in CNNs are often not obvious to be interpreted \cite{Zeiler2011Adaptive}. 

On the other hand, researchers begin to adapt CNNs for studying the target questions from neuroscience. For example, CNNs have been used to model the ventral visual pathway that has been suggested as a route for visual object recognition starting from the retina to visual cortex and reaching inferior temporal (IT) cortex \cite{Yamins2013Hierarchical, Yamins_2014, Cadieu2014Deep}. The prediction of neuronal responses recorded in monkey in this case has a surprisingly good performance. However, the final output of this CNN model is representing dense computations conducted in many previous conv layers, which may or may not be related to the neuronscience underpinnings of information processing in the brain. Understanding these network components of CNN are difficult given the IT cortex part is sitting at a high level of our visual system with abstract information, if any, encoded \cite{Yamins_2016}. In principle, CNN models can also be applied to early sensory systems where the organization of underlying neuronal circuitry is relatively more clear and simple. Thus one expect knowledge of these neuronal circuitry could provide useful and important validation for such models. For instance, a recent study employs CNNs to predict neural responses of the retinal ganglion cells to white noise and natural images \cite{McIntosh2016}. 

Here we move a step further in this direction by relating CNNs with single RGCs. Specifically, we used CNNs to learn to predict the responses of single RGCs to white noise images. In contrast to the study by \cite{McIntosh2016} where one single CNN model was used to model a population of RGCs, in the current study, our main focus is based on single RGCs to revealing the network structure components learned by CNNs. Our aim is to study what kind of possible biological structure components in the retina can be learned by CNNs. This concerns the research focus of understanding, visualizing and interpreting the CNN components out of its black box.  

To the end, by using a minimal model of RGC, we found the conv filters learned in CNN are essentially the subunit components of RGC model. The features represented by these filters are fallen into the receptive field of modeled RGC. Furthermore, we applied CNNs to analyze biological RGC data recorded in salamander. Surprisingly, the conv filters are resembling to the receptive fields of bipolar cells that sit in the previous layer of RGC and pool their computations to a downstream single RGC.    

\section{Methods}

\subsection{RGC model and data}

A simulated RGC is modeled in Fig.~\ref{fig:01} as previously \cite{Liu2017}. The model cell has five subunits that each filter, similar to a conv filter in DNN, convolves the incoming stimulus image and then applies an nonlinearity of threshold-quadratic rectification. The subunit signals are then polled together by the RGC. The polled signal is applied with a threshold-linear output nonlinearity with a positive threshold at unity to make spiking sparse. 

The biological data of RGCs were recorded in salamander as described in \cite{Liu2017}. Briefly RGC spiking activity were obtained by multielectrode array recordings as in \cite{Liu_2015}. The retinas was optically stimulated with spatiotemporal white noise images, temporally updated at a rate of 30 Hz and spatially arranged in a checkerboard layout with stimulus pixels of 30x30 $m\mu$. 

\subsection{CNN model}

We adopt a CNN model containing two convolution layers and a dense layer as in \cite{McIntosh2016}. Several sets  of parameters in convolution layers, including the number of layers, the number and size of convolution filters were explored. The predication performance is robust against these changes of parameters. Therefore we used a filter size as of $15\times15$ to compare our results with those in \cite{McIntosh2016}. The major difference between our model with that in \cite{McIntosh2016} is that our CNN is for studying of single RGCs.

For the RGC model, we used a data set consisting of 600k training samples of white noise images, and additional set of samples for testing. The training labels are a train of binary spikes with 0 and 1 generated by the model. For the biological RGCs recorded in salamander, we used the same data sets as in \cite{Liu2017}. Briefly there are about 40k training samples and labels with the number of spikes as in [0 5] for each image. The test data have 300 samples, which are repeatedly presented to the retina for about 200 trials. The average firing rate of this test data is compared to the CNN output for performance calculation.

\section{Results}

Here we focus on single RGC that has the benefit to clarify the network structure components of CNNs. Recently, a variation of non-negative matrix factorization was used to analyze the RGC' responses to white noise images and identify a number of subunits, resembling to the biopolar cells, within the receptive field of each RGC \cite{Liu2017}. With this picture in mind, here we address the question that what types of network structure components can be revealed by CNNs when they are used to model the single RGC response.

A previous study \cite{McIntosh2016} focused on predicating neural response of RGC at the population level with one CNN model, and claimed that the features represented by conv filters are resembling to the receptive fields of bipolar cells (BCs). However, a careful examination reveals that this connection between CNN feature map and BCs is weak since the number of conv filters in the CNN is much less than that of BCs in the RGC population from the retina. By using a CNN model, one expect to reveal a more clear picture of this connection between CNNs and the retina.

\begin{figure*}[t]
	\begin{center}
		\includegraphics[width=13cm]{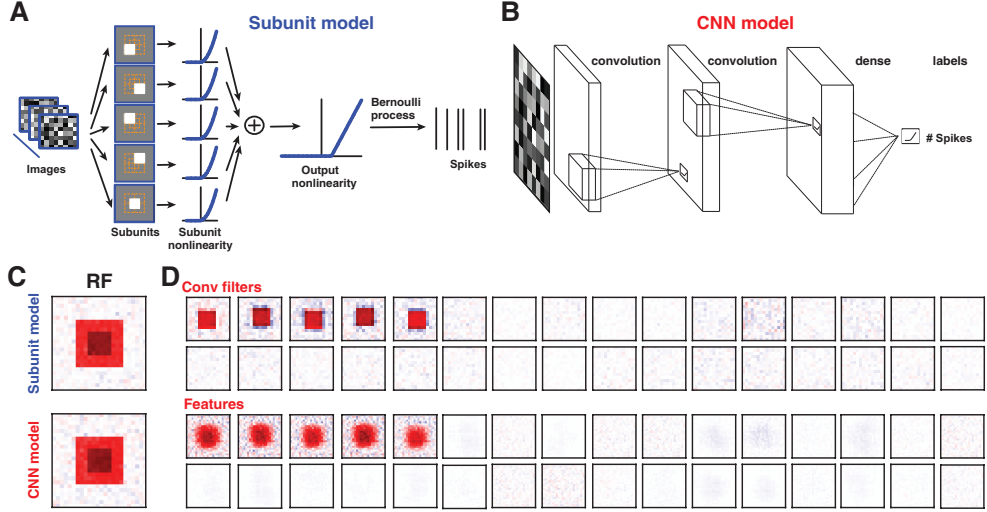}
	\end{center}
	\caption{The CNN filters are resembling to the subunits in RGC model. (A) An illustration of RGC model structure. Note there are five subunits playing the role of conv filters. (B) An illustration of CNN model that trains the same set of images to predicate the labels, here spikes, of all images. (C) Receptive fields (RFs) of modeled RGC and predication by CNN model. (D) Visualizing of CNN model components of both conv filters and average features represented by each filter. }
	\label{fig:01}
\end{figure*}

Here, we set up a single RGC model with conv subunits as in Fig.~\ref{fig:01}(A), which is resembling to a 2-layer CNN with one conv layer of subunits and one dense layer of single RGC. By training a CNN as in Fig.~\ref{fig:01}(B) with a set of white noise images to predicate the target labels as the simulated spikes generated by this RGC model, we found that the CNN model can predicate the RGC model response well with Pearson correlation coefficient (CC) up to 0.70 similar to the study by \cite{McIntosh2016}. 

Interestingly, we also found the CNN model can predicate the receptive field (RF) well as in Fig.~\ref{fig:01}(C). Furtherer more, the conv filters learned by CNN are the \textmd{exact} subunits employed in the RGC model as shown in Fig.~\ref{fig:01}(C). A subset of the conv filters, that can be termed as \textrm{effective} filters, start from random shapes and converge to the exact subunits. Although we set up the filter size as 15x15 pixels, the resulting effective filters are sparse represented with a 6x6 pixel size. The rest of the filters are still random and close to zero. Therefore, these results show that CNN parameters are highly redundant. Such a redundancy of parameters, including conv filters, units/neurons and connections of conv and dense layers, is widely observed for deep learning models \cite{Denil_2013,Han2015Deep,Han_NIPS2015}. All together, These results suggest that the CNN model can identify the underlying hidden network structure components within the RGC model by only looking at the input stimulus images and the output response in terms of the number of spikes.  

\begin{figure*}[!tbh]
	\begin{center}
		\includegraphics[width=13cm]{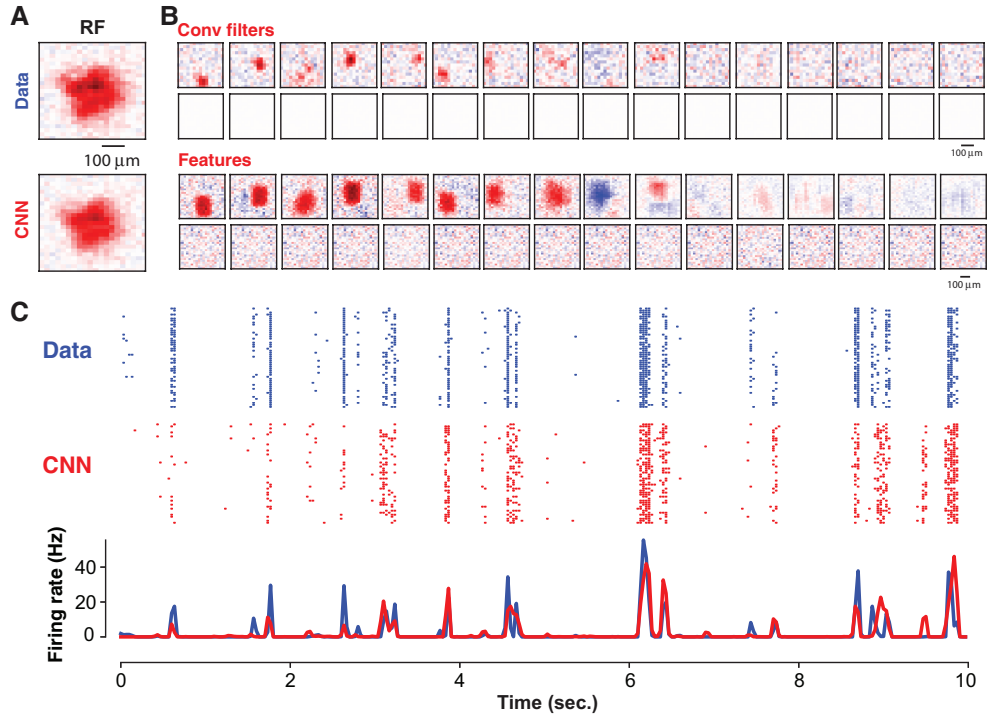}
	\end{center}
	\caption{The CNN reveals subunit structures in biological RGC data. (A) Receptive fields of the sample cell and CNN predication. (B) Visualizing of CNN model components of both conv filters and average features represented by each filter. (C) Neural response predicated well by CNN model visualized by RGC data spike rasters (upper) and CNN spike rasters (middle) and their average firing rates. }
	\label{fig:02}
\end{figure*}

To further characterizing these structure components in details, we use a CNN to learn the biological RGC data with the similar images of white noise and the spiking responses. Similar to the results of the RGC model above, the outputs of CNN model can recover the receptive field of data very well as in Fig.~\ref{fig:02}(A). We also found that the learned conv filters converge to a set of localized subunits whereas the rest of filters are noisy and close to zero as in Fig.~\ref{fig:02}(B). The size of these localized filters is comparable to that in bipolar cells around 100 $m\mu$ \cite{Liu2017}. 

In addition, the features represented by these localized conv filers are also localized. Given the example RGC is a OFF type cell that response to the dark part of images strongly, most features have similar OFF peaks resulted from the OFF BC-like filters. These OFF features tile the space of receptive field of RGC. Interestingly, there are some features with ON peaks, which play a role as inhibition in the retinal circuit. A few features have some complex structures mixed with OFF and ON peaks, which are mostly resulted from the less localized filters. However, if the filters are pure noise, the resulting features are pure noise without any structure embedded. Besides filters and features, the CNN model generates a good predication of RGC response as in Fig.~\ref{fig:02}(C) with the CC up to 0.75. These observations are similar across different RGCs recorded.

\section{Discussion}
Here by focusing on single RGCs, we shown that CNN can learn their parameters in an interpretable fashion. Both filters and features are close to the biological underpinnings within the retinal circuit. With the benefits of relative well-understood neuronal circuit of the retina ganglion cells, our preliminary results give a strong evidence that the building-blocks of CNNs are meaningful when they are applied to neuroscience for revealing network structure components. Our results extend the previous studies \cite{Yamins_2014,McIntosh2016} that focus on predication of neural responses. Furtherer more, our approach is suitable to address other difficult issues of deep learning, such as transfer learning, since the domain of images seen by single RGCs is local and less complicated than those global structures of entire natural images.

\section{Acknowledgements}
Q.Y., Z.Y. and F.C. were supported in part by the National Natural Science Foundation
of China under Grant 61671266, 61327902, in part by the Research Project of Tsinghua University under Grant 20161080084, in part by National
High-tech Research and Development Plan under Grant 2015AA042306. J.K.L. was partially supported by the Human Brain Project of the European Union \#604102 and \#720270.

%
%
\small
\bibliography{ref}

\begin{thebibliography}{10}

\bibitem{Lecun2015Deep}
Yann Lecun, Yoshua Bengio, and Geoffrey Hinton.
\newblock Deep learning.
\newblock {\em Nature}, 521(7553):436--444, 2015.

\bibitem{Smith_2016}
Leslie~N. Smith and Nicholay Topin.
\newblock Deep convolutional neural network design patterns.
\newblock {\em arXiv:1611.00847v3}, 2016.

\bibitem{Marblestone_2016}
Adam~H. Marblestone, Greg Wayne, and Konrad~P. Kording.
\newblock Toward an integration of deep learning and neuroscience.
\newblock {\em Frontiers in Computational Neuroscience}, 10, sep 2016.

\bibitem{Zeiler2014Visualizing}
Matthew~D. Zeiler and Rob Fergus.
\newblock Visualizing and understanding convolutional networks.
\newblock In {\em European Conference on Computer Vision}, pages 818--833,
  2014.

\bibitem{Lecun2010Convolutional}
Y~Lecun, K~Kavukcuoglu, and C~Farabet.
\newblock Convolutional networks and applications in vision.
\newblock In {\em IEEE International Symposium on Circuits and Systems}, pages
  253--256, 2010.

\bibitem{Simonyan2014Very}
Karen Simonyan and Andrew Zisserman.
\newblock Very deep convolutional networks for large-scale image recognition.
\newblock {\em Computer Science}, 2014.

\bibitem{Krizhevsky2012ImageNet}
Alex Krizhevsky, Ilya Sutskever, and Geoffrey~E. Hinton.
\newblock Imagenet classification with deep convolutional neural networks.
\newblock In {\em International Conference on Neural Information Processing
  Systems}, pages 1097--1105, 2012.

\bibitem{Simoncelli2001Natural}
E.~P. Simoncelli and B.~A. Olshausen.
\newblock Natural image statistics and neural representation.
\newblock {\em Annual Review of Neuroscience}, 24(24):1193, 2001.

\bibitem{Zeiler2011Adaptive}
Matthew~D. Zeiler, Graham~W. Taylor, and Rob Fergus.
\newblock Adaptive deconvolutional networks for mid and high level feature
  learning.
\newblock In {\em International Conference on Computer Vision}, pages
  2018--2025, 2011.

\bibitem{Yamins2013Hierarchical}
Daniel Yamins, Ha~Hong, Charles Cadieu, and James~J. Dicarlo.
\newblock Hierarchical modular optimization of convolutional networks achieves
  representations similar to macaque it and human ventral stream.
\newblock {\em Advances in Neural Information Processing Systems}, pages
  3093--3101, 2013.

\bibitem{Yamins_2014}
D.~L.~K. Yamins, H.~Hong, C.~F. Cadieu, E.~A. Solomon, D.~Seibert, and J.~J.
  DiCarlo.
\newblock Performance-optimized hierarchical models predict neural responses in
  higher visual cortex.
\newblock {\em Proceedings of the National Academy of Sciences},
  111(23):8619--8624, may 2014.

\bibitem{Cadieu2014Deep}
Charles~F. Cadieu, Ha~Hong, Daniel L.~K. Yamins, Nicolas Pinto, Diego Ardila,
  Ethan~A. Solomon, Najib~J. Majaj, and James~J. Dicarlo.
\newblock Deep neural networks rival the representation of primate it cortex
  for core visual object recognition.
\newblock {\em Plos Computational Biology}, 10(12):e1003963, 2014.

\bibitem{Yamins_2016}
Daniel L~K Yamins and James~J DiCarlo.
\newblock Using goal-driven deep learning models to understand sensory cortex.
\newblock {\em Nature Neuroscience}, 19(3):356--365, feb 2016.

\bibitem{McIntosh2016}
Lane McIntosh, Niru Maheswaranathan, Aran Nayebi, Surya Ganguli, and Stephen
  Baccus.
\newblock Deep learning models of the retinal response to natural scenes.
\newblock In {\em Advances in Neural Information Processing Systems 29}. 2016.

\bibitem{Liu2017}
Jian~K. Liu, Helene~M. Schreyer, Arno Onken, Fernando Rozenblit, Mohammad~H.
  Khani, Vidhyasankar Krishnamoorthy, Stefano Panzeri, and Tim Gollisch.
\newblock Inference of neuronal functional circuitry with spike-triggered
  non-negative matrix factorization.
\newblock {\em Nature Communications}, 8(1), jul 2017.

\bibitem{Liu_2015}
Jian~K. Liu and Tim Gollisch.
\newblock Spike-triggered covariance analysis reveals phenomenological
  diversity of contrast adaptation in the retina.
\newblock {\em {PLOS} Computational Biology}, 11(7):e1004425, jul 2015.

\bibitem{Denil_2013}
Misha Denil, Babak Shakibi, Laurent Dinh, MarcAurelio Ranzato, and Nando
  de~Freitas.
\newblock Predicting parameters in deep learning.
\newblock In C.~J.~C. Burges, L.~Bottou, M.~Welling, Z.~Ghahramani, and K.~Q.
  Weinberger, editors, {\em Advances in Neural Information Processing Systems
  26}, pages 2148--2156. Curran Associates, Inc., 2013.

\bibitem{Han2015Deep}
Song Han, Huizi Mao, and William~J Dally.
\newblock Deep compression: Compressing deep neural networks with pruning,
  trained quantization and huffman coding.
\newblock {\em Fiber}, 56(4):3--7, 2015.

\bibitem{Han_NIPS2015}
Song Han, Jeff Pool, John Tran, and William Dally.
\newblock Learning both weights and connections for efficient neural network.
\newblock In C.~Cortes, N.~D. Lawrence, D.~D. Lee, M.~Sugiyama, and R.~Garnett,
  editors, {\em Advances in Neural Information Processing Systems 28}, pages
  1135--1143. Curran Associates, Inc., 2015.

\end{thebibliography}

\end{document}